\documentclass[12pt]{article}

\usepackage[letterpaper,left=1.25in,right=1.25in,top=1in,bottom=1in]{geometry}
\usepackage[T1]{fontenc}
\usepackage[utf8]{inputenc}
\usepackage{newtxtext,newtxmath}
\usepackage{setspace}
\usepackage{titlesec}
\titleformat{\section}{\bfseries\fontsize{16}{18}\selectfont}{\thesection}{1em}{}
\titlespacing*{\section}{0pt}{2\baselineskip}{1\baselineskip}
\titleformat{\subsection}{\bfseries\fontsize{14}{16}\selectfont}{\thesubsection}{1em}{}
\titlespacing*{\subsection}{0pt}{2\baselineskip}{1\baselineskip}
\titleformat{\subsubsection}{\bfseries\fontsize{12}{14}\selectfont}{\thesubsubsection}{1em}{}
\titlespacing*{\subsubsection}{0pt}{2\baselineskip}{1\baselineskip}
\usepackage{fancyhdr}
\usepackage{graphicx}
\usepackage{float}
\usepackage{booktabs}
\usepackage{amsmath}
\usepackage{caption}
\usepackage[numbers]{natbib}
\usepackage[colorlinks=true,linkcolor=black,citecolor=black,urlcolor=black]{hyperref}

\newcommand{\MICSTitle}[1]{%
  \begin{center}
    \vspace*{1.5in}
    {\fontsize{18}{20}\selectfont #1\par}
    \vspace{2\baselineskip}
  \end{center}
}

\newcommand{\MICSAuthorBlock}[5]{%
  \begin{center}
    {\fontsize{14}{16}\selectfont #1\par}
    {\fontsize{14}{16}\selectfont #2\par}
    {\fontsize{14}{16}\selectfont #3\par}
    {\fontsize{14}{16}\selectfont #4\par}
    {\fontsize{14}{16}\selectfont #5\par}
    \vspace{2\baselineskip}
  \end{center}
}

\newcommand{\MICSAbstract}[1]{%
  \begin{center}
    {\bfseries\fontsize{16}{18}\selectfont Abstract\par}
  \end{center}
  \vspace{\baselineskip}
  {\fontsize{12}{14}\selectfont #1\par}
}

\begin{document}

\thispagestyle{empty}

\MICSTitle{Source-Aware Reranking for Retrieval-Augmented Generation:\\
A Reliability Prior Approach}

\MICSAuthorBlock
  {Yuktha Tata Koganti, Hugo Garrido-Lestache Belinchon}
  {Department of Computer Science and Software Engineering}
  {Milwaukee School of Engineering}
  {1025 N Broadway St, Milwaukee, WI 53202}
  {kogantiy@msoe.edu, garrido-lestacheh@msoe.edu}

\MICSAbstract{%
Standard Retrieval-Augmented Generation pipelines rank retrieved documents
by semantic similarity alone, without accounting for source provenance or
credibility. This work evaluates a simple and interpretable modification to
RAG retrieval ranking that incorporates domain-informed source reliability
priors. Each document is assigned a prior $\lambda(s)$ based on its source
type, and retrieval scores are reweighted using
$\text{score}(q,d) = \text{sim}(q,d) \times \lambda(s)$. The framework is
evaluated against a similarity-only baseline on a 120-document health-domain
corpus. In this controlled setting, source-aware reranking improves
Precision@5 from 0.48 to 0.72 and reduces average adversarial document retrieval
under the evaluated threat model, where low-credibility sources are
identifiable via metadata. All experiments were executed on Rosie, the
high-performance computing cluster at the Milwaukee School of Engineering,
which provided the GPU-accelerated infrastructure necessary to run the full
experimental pipeline reliably and reproducibly. These results suggest a
potential mitigation strategy for source quality degradation in RAG
pipelines, within the limits of the experimental setup described.
}

\newpage
\setcounter{page}{1}
\pagestyle{fancy}

\section{Introduction}

Large language models are increasingly deployed in domains where factual
accuracy carries real-world consequences, including medicine, law, and public
health. Retrieval-Augmented Generation addresses a fundamental limitation of
static language models by grounding responses in externally retrieved
documents, allowing systems to incorporate up-to-date and domain-specific
knowledge at inference time. However, standard RAG pipelines retrieve
documents based solely on semantic similarity between query and document
embeddings. This design treats all retrieved documents as equally valid inputs
regardless of their origin, credibility, or potential for harm.

In open-domain or uncurated deployment settings, this can introduce a
potential vulnerability. A peer-reviewed clinical guideline and an
AI-generated post may produce similar embedding representations if their
surface content addresses the same topic, and a similarity-only retriever has
no signal to distinguish between them based on source credibility. In
adversarial settings, semantically plausible but factually misleading
documents can be injected into a retrieval corpus and retrieved without
penalty under a similarity-only ranking function.

This work addresses the problem by introducing a source reliability prior
into the RAG retrieval ranking function. The framework assigns each document
a reliability prior based on its source type, multiplies retrieval similarity
scores by this prior during reranking, and optionally propagates reliability
signals to semantically adjacent documents through a single-hop smoothing
step. A post-query weight update mechanism adjusts priors based on heuristic
feedback across sequential interactions.

This work does not propose a new retrieval model. It evaluates a simple and
interpretable modification to retrieval ranking that incorporates
source-level priors, and examines whether this modification reduces retrieval
of low-credibility content in a controlled setting. The contribution lies in
systematically characterizing how source-level priors interact with retrieval
behavior across standard, adversarial, and parameter sensitivity settings.

This paper makes three contributions. First, a reliability-weighted
reranking formulation that operates as a post-retrieval modification without
altering the underlying language model or embedding model. Second, a
controlled adversarial injection evaluation under a specific threat model
where low-credibility documents are identifiable by source type
metadata---results are interpreted within this threat model only. Third, an
empirical analysis on a 120-document health-domain corpus suggesting that
source provenance weighting, in this controlled setting, reduces adversarial
retrieval and improves precision over a similarity-only baseline.

\section{Related Work}

\textbf{Retrieval-Augmented Generation.}
Lewis et al.\ introduced RAG as a framework combining parametric language
model knowledge with non-parametric dense retrieval over document
collections, demonstrating substantially reduced hallucination in open-domain
question answering~\cite{lewis2020rag}. Guu et al.\ extended this direction
with REALM, which jointly trains the retriever and generator
end-to-end~\cite{guu2020realm}. Both systems retrieve documents purely on
the basis of embedding similarity and include no mechanism for assessing
source credibility. Our work extends this foundation by introducing a
multiplicative reliability term into the retrieval ranking function.

\textbf{Dense Passage Retrieval and Indexing.}
Karpukhin et al.\ established dense passage retrieval using bi-encoder
architectures as an effective alternative to sparse retrieval
methods~\cite{karpukhin2020dpr}. Johnson et al.\ developed FAISS as a
scalable approximate nearest-neighbor index enabling efficient similarity
search over large document collections~\cite{johnson2019faiss}. Our
retrieval layer builds directly on these methods, using SentenceTransformers
with FAISS for candidate retrieval before applying reliability reranking.

\textbf{Adversarial Robustness in Neural Retrieval.}
Prior work has demonstrated that neural retrieval systems are susceptible to
adversarial passages crafted to appear semantically relevant while containing
misleading content. Carlini et al.\ and related work on corpus poisoning have
shown that embedding-based retrievers can be manipulated through injection of
high-similarity low-quality documents~\cite{carlini2024poisoning}. Our
adversarial injection experiment directly operationalizes this threat model
and measures the suppression effect of reliability priors under specific and
bounded conditions.

\textbf{Trust and Provenance in Information Systems.}
Research in information credibility and knowledge graph quality has
established that source provenance is a meaningful signal for assessing
reliability~\cite{dong2014knowledge}. Work on misinformation detection has
demonstrated that source-level signals provide predictive value independent
of content-level signals~\cite{perezrosas2018fakenews}. Our work differs
from these approaches in that it operates at the retrieval layer of an LLM
pipeline rather than as a standalone classification system, and uses
predefined domain-informed priors rather than learned classifiers.

\section{Implementation}

\subsection{System Architecture}

The proposed framework extends the standard RAG retrieval pipeline with a
source-aware reranking layer that operates between the FAISS index and the
final context window. The system shares its embedding and retrieval
components with the baseline: documents are embedded using a sentence
transformer encoder, indexed via approximate nearest-neighbor search, and
retrieved as a candidate pool. The source-aware system then applies
reliability weighting, optional neighbor smoothing, and optional post-query
weight updates before returning the final top-$k$ results. The baseline
returns top-$k$ results directly from FAISS by similarity rank. Both systems
are evaluated using identical metrics on identical query sets.

The core ranking modification---multiplying similarity by a source
prior---is intentionally simple. The contribution of this work is not
algorithmic novelty but an empirical evaluation of whether this simple
modification produces meaningful retrieval differences in a controlled
setting, and an honest characterization of its limitations and boundary
conditions.

\vspace{-0.5\baselineskip}
\begin{figure}[H]
  \centering
  \includegraphics[width=0.95\linewidth]{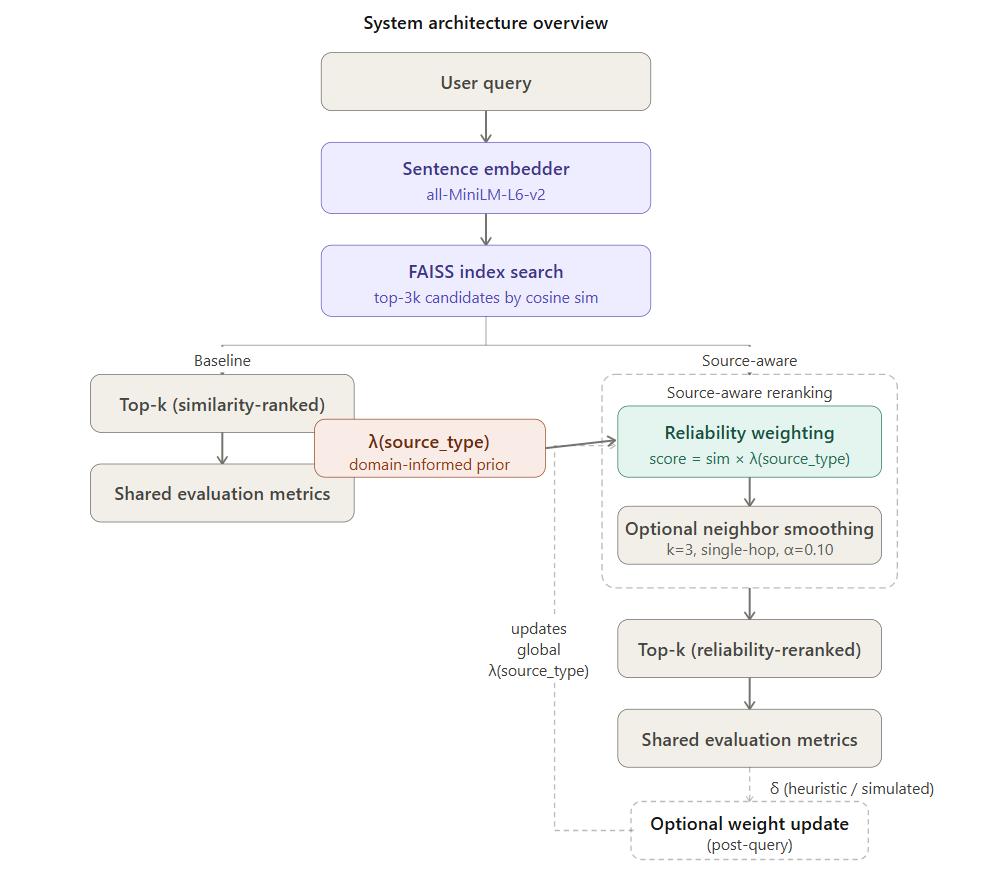}
  \caption{System architecture overview. The left branch shows the
  baseline similarity-only pipeline. The right branch shows the
  source-aware reranking pipeline with reliability weighting, optional
  neighbor smoothing, and optional post-query weight update. Both branches
  share embedding and retrieval components and are evaluated on identical
  query sets.}
  \label{fig:architecture}
\end{figure}

\subsection{Corpus and Source Taxonomy}

The evaluation corpus consists of 120 documents in the health and medical
information domain, spanning seven source types. The health domain was
selected because source reliability carries clear real-world meaning in this
context---the distinction between a peer-reviewed clinical study and an
AI-generated health post carries practical significance for downstream answer
quality. The corpus includes 10 adversarially crafted documents with source
types \texttt{ai\_generated} or \texttt{user\_generated}, each written to be
semantically close to the evaluation query topics while containing misleading
or false medical claims. These documents are known only to the evaluation
layer and are not flagged during retrieval.

\subsection{Baseline Retrieval}

The baseline system embeds all documents using the all-MiniLM-L6-v2 sentence
transformer~\cite{reimers2019sbert} and indexes them using FAISS with inner
product similarity over L2-normalized vectors, equivalent to cosine
similarity. For a query $q$ and document $d$, the baseline retrieval score
is:

\begin{equation}
  \text{score}_{\text{baseline}}(q, d) = \text{sim}(q, d)
\end{equation}

The top-$k$ documents by this score form the baseline retrieved set with no
further modification.

\subsection{Reliability-Weighted Reranking}

The source-aware system retrieves a candidate pool of $3k$ documents by
similarity, then reranks using a reliability-weighted score:

\begin{equation}
  \text{score}_{\text{SA}}(q, d) = \text{sim}(q, d) \times \lambda(s_d)
\end{equation}

where $s_d$ is the source type of document $d$ and $\lambda(s_d)$ is a
predefined domain-informed prior. The expanded candidate pool of $3k$ is
necessary to allow reliability weighting to meaningfully reorder
results---reranking only the final top-$k$ would produce no change in
ranking order. Initial priors are shown in Table~\ref{tab:priors}.

\begin{table}[H]
\centering
\caption{Source type reliability priors $\lambda(s)$.}
\begin{tabular}{lc}
\toprule
Source Type & $\lambda(s)$ \\
\midrule
peer\_reviewed   & 0.95 \\
government       & 0.90 \\
established\_news & 0.75 \\
organization\_report & 0.70 \\
blog             & 0.50 \\
user\_generated  & 0.40 \\
ai\_generated    & 0.30 \\
unknown          & 0.35 \\
\bottomrule
\end{tabular}
\label{tab:priors}
\end{table}

These values represent domain-informed assumptions about relative source
credibility in the health domain. They are not learned from data and are not
claimed to be optimal. The sensitivity of results to these values is examined
empirically in Section~\ref{sec:sensitivity}.

\subsection{Optional Neighbor Smoothing}

After reliability weighting, an optional single-hop smoothing step adjusts
each candidate document's effective reliability based on its nearest
neighbors in the index:

\begin{equation}
  \lambda_{\text{smoothed}}(d) = \lambda(d)
    + \alpha \sum_{n \in \mathcal{N}(d)} \text{sim}(d,n) \cdot \lambda(n)
\end{equation}

where $\mathcal{N}(d)$ denotes the top-3 nearest neighbors of $d$,
$\alpha = 0.10$ is a fixed smoothing coefficient, and the result is clipped
to $[0.10,\, 1.00]$. This step is a heuristic adjustment, not a principled
propagation algorithm. On the current corpus its effect is marginal; it is
included as a structural extension whose benefit may increase with corpus
size.

\subsection{Optional Post-Query Weight Update}

After each query, the system optionally updates source type weights using a
heuristic feedback signal:

\begin{equation}
  \lambda_{t+1}(s) =
    \text{clip}\!\left(\lambda_t(s) + \eta\,\delta_t,\; 0.10,\; 1.00\right)
\end{equation}

where $\eta = 0.05$ is the update rate and $\delta_t$ is a simulated
feedback signal based on ground-truth relevance labels. Specifically:
$\delta = {+}0.3$ if the source type matches expected relevant types,
$\delta = {-}0.5$ if the document is adversarial, and $\delta = {-}0.1$
otherwise. This feedback is oracle-based---it uses ground-truth labels
rather than real user signals---and results from this component should be
interpreted as demonstrating the behavior of the update rule under ideal
conditions, not as evidence of deployable adaptive learning.

\subsection{Computational Infrastructure}

All experiments in this work were executed and validated on Rosie, the
high-performance computing cluster operated by the Milwaukee School of
Engineering. The repeated embedding and indexing workload across baseline,
source-aware, adversarial, sensitivity, and adaptive experimental runs would
have been prohibitively slow on standard laptop hardware. Rosie's
GPU-accelerated nodes allowed the full pipeline to complete in a single
session with consistent, reproducible results. Its stable, dependency-isolated
Linux environment ensured identical software conditions across all runs, and
version control via Git enabled every reported result to be traced to a
specific experimental configuration. The breadth of evaluation in this paper
was only practical because of the infrastructure Rosie provided.

\section{Experimentation}

\subsection{Evaluation Setup}

Ten evaluation queries were constructed across five medical topics: vaccine
safety, drug interactions, cancer screening, antibiotic resistance, and
treatment efficacy. Each query is paired with a set of relevant document IDs
based on topical match, defined independently of source type. Retrieval is
evaluated at $k=5$ for all systems using the following metrics: Precision@5
(proportion of top-5 results that are relevant), NDCG@5 (normalized
discounted cumulative gain using relevance labels as grades), MRR (reciprocal
rank of the first relevant document), and adversarial hit rate (proportion of
top-5 results that are adversarial documents). Average reliability score and
high-reliability ratio are reported separately as secondary descriptive
metrics directly tied to the method's design.

\subsection{Adversarial Injection Protocol}

All 10 adversarial documents were injected into the FAISS index alongside
the 110 clean documents for the adversarial experiment. Both systems were
queried over all 10 evaluation queries with the full poisoned index. The
adversarial documents were designed to be semantically close to query topics,
representing a hard case for similarity-based retrieval. The adversarial hit
rate measures how frequently these documents appear in the final top-5
results.

\subsection{Adaptive Learning Protocol}

The adaptive experiment processes all 10 queries sequentially, updating
weights after each query using the heuristic signal described in
Section~3.6. Weight history is logged after every query. Convergence is
assessed as mean absolute weight change over the final three queries falling
below 0.01.

\section{Results and Analysis}

\subsection{Retrieval Quality}

Table~\ref{tab:retrieval} presents average retrieval metrics across all 10
queries for both systems.

\begin{table}[H]
\centering
\caption{Retrieval metrics averaged across 10 queries,
         baseline vs.\ source-aware.}
\begin{tabular}{lccc}
\toprule
Metric & Baseline & Source-Aware & Change \\
\midrule
Precision@5 & 0.48 & 0.72 & $+$50.0\% \\
NDCG@5      & 0.907 & 0.976 & $+$7.6\% \\
MRR         & 0.833 & 1.000 & $+$20.0\% \\
\bottomrule
\end{tabular}
\label{tab:retrieval}
\end{table}

In this controlled setting, source-aware reranking improves Precision@5 from
0.48 to 0.72, a 50\% relative improvement, indicating that reliability
weighting consistently surfaces topically relevant documents. The
source-aware system achieves MRR of 1.000 across all queries, meaning the
first retrieved document is always relevant across this query set. The
consistently perfect MRR suggests that the evaluation queries are relatively
well-aligned with relevant documents in the corpus, which may limit the
ability to differentiate systems on this metric.
Precision@5 across individual queries ranged from 0.40 to 1.00 for the
source-aware system and from 0.40 to 0.80 for the baseline, indicating
consistent directional improvement across queries though the magnitude
varies. No formal significance testing was performed given the query count
of 10, and these results should be treated as exploratory rather than
statistically confirmed.

\begin{figure}[H]
  \centering
  \includegraphics[width=0.95\linewidth]{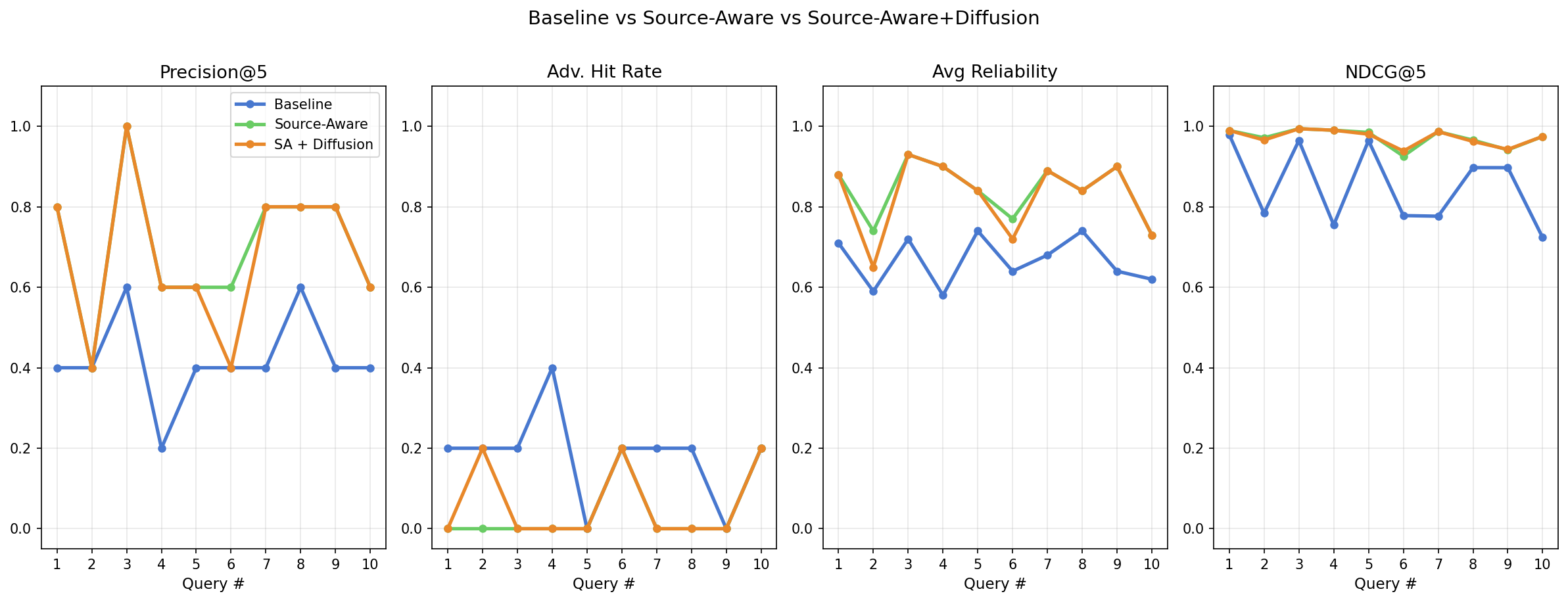}
  \caption{Comparison of Baseline, Source-Aware, and Source-Aware with neighbor
  diffusion across Precision@5, adversarial hit rate, average reliability, and
  NDCG@5 per query. SA+Diffusion consistently matches or exceeds Source-Aware
  on reliability and adversarial suppression across most queries.}
  \label{fig:diffusion}
\end{figure}

Secondary reliability metrics are reported for descriptive purposes: the
source-aware system achieved an average reliability score of 0.863 versus
0.737 for the baseline. These metrics are directly tied to the method's
design and do not constitute independent evidence of retrieval quality
improvement. Figure~\ref{fig:diffusion} further shows that adding neighbor
diffusion on top of source-aware reranking consistently matches or exceeds
the source-aware system on reliability and adversarial suppression across
most queries.

\begin{figure}[H]
  \centering
  \includegraphics[width=0.95\linewidth]{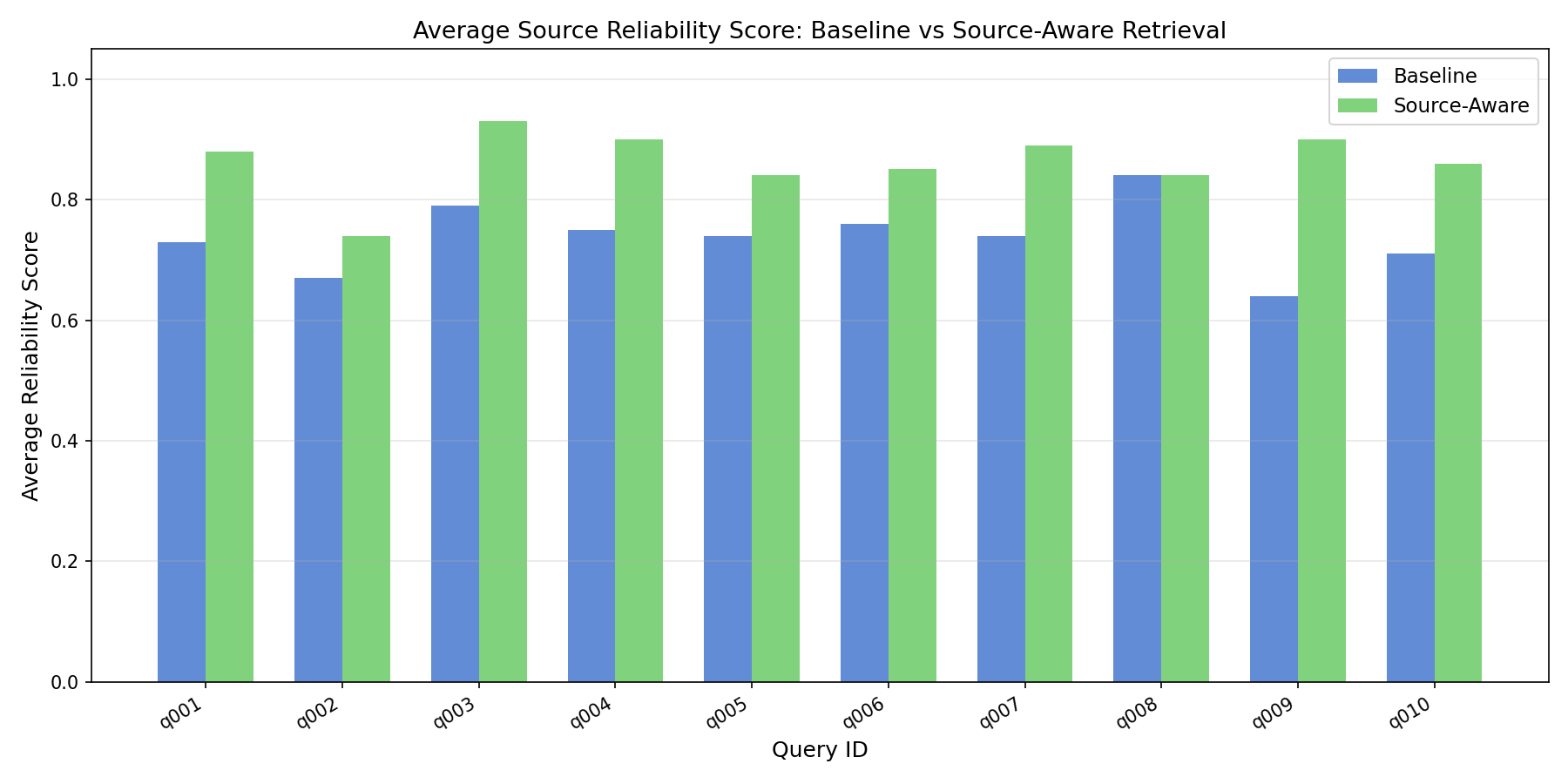}
  \caption{Average source reliability score per query, baseline
           vs.\ source-aware. Reported as a secondary descriptive metric.}
  \label{fig:reliability}
\end{figure}

\begin{figure}[H]
  \centering
  \includegraphics[width=0.95\linewidth]{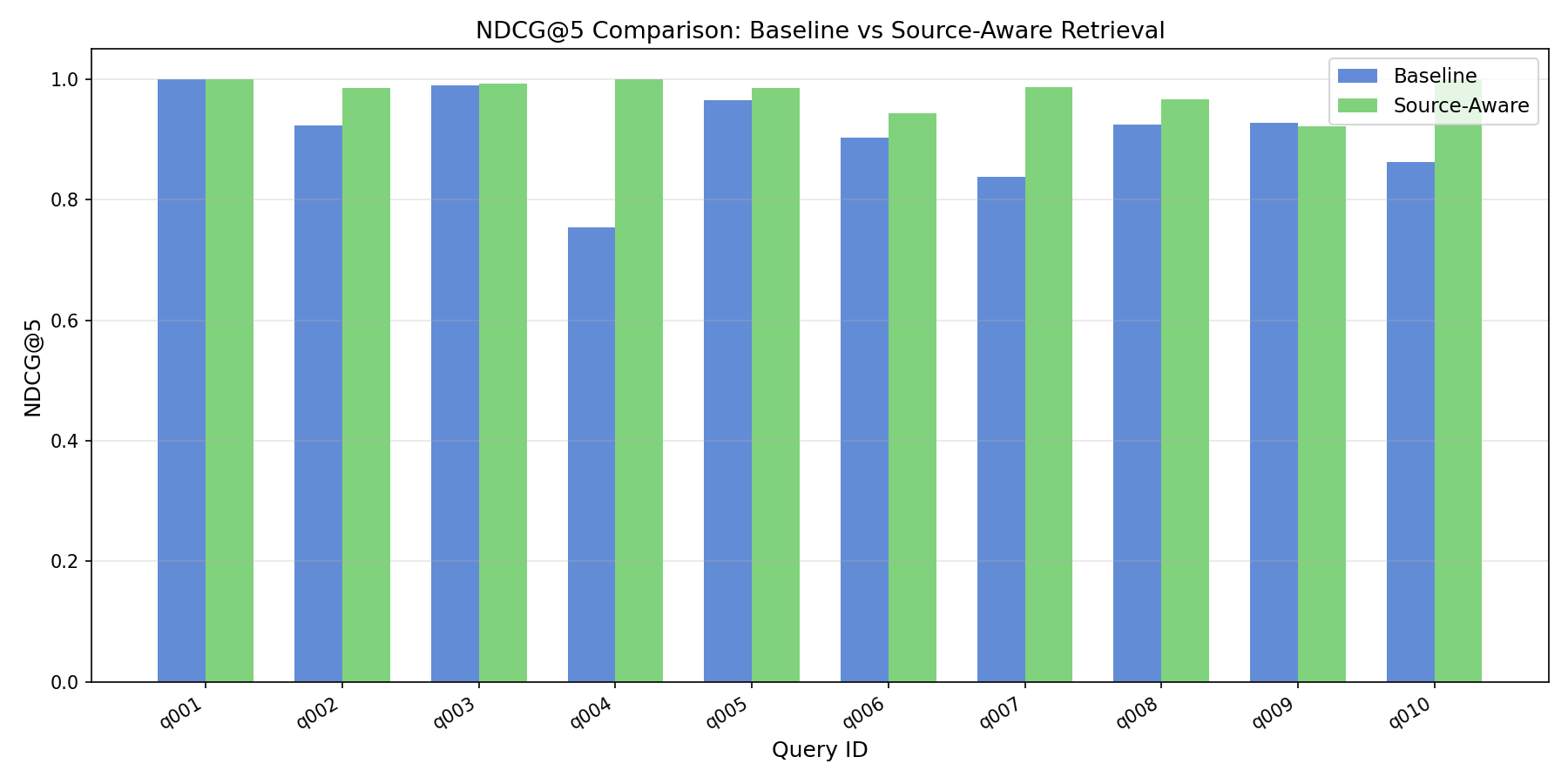}
  \caption{NDCG@5 per query, baseline vs.\ source-aware retrieval.}
  \label{fig:ndcg}
\end{figure}

\subsection{Adversarial Robustness}

Table~\ref{tab:adversarial} presents adversarial hit rates under the
injection protocol. The baseline retrieves at least one adversarial document
in 9 of 10 queries, with an average adversarial hit rate of 0.32.
Source-aware reranking reduces this to an average of 0.28, a 12.5\%
relative reduction. In 6 of 10 queries, source-aware retrieval either
eliminates or ties adversarial exposure. However, two queries (q006 and
q010) showed a slight increase in adversarial hit rate under the source-aware
system, indicating that reliability weighting is not uniformly beneficial
and can in some cases surface adversarial documents that rank higher after
reweighting displaces other low-reliability clean documents.

\begin{table}[H]
\centering
\caption{Adversarial document hit rate per query topic.}
\begin{tabular}{lcc}
\toprule
Topic & Baseline & Source-Aware \\
\midrule
vaccine\_safety (q001)        & 0.40 & 0.20 \\
vaccine\_safety (q002)        & 0.40 & 0.20 \\
drug\_interaction (q003)      & 0.40 & 0.40 \\
drug\_interaction (q004)      & 0.40 & 0.40 \\
cancer\_screening (q005)      & 0.20 & 0.20 \\
cancer\_screening (q006)      & 0.20 & 0.40 \\
antibiotic\_resistance (q007) & 0.40 & 0.20 \\
antibiotic\_resistance (q008) & 0.40 & 0.20 \\
treatment\_efficacy (q009)    & 0.20 & 0.20 \\
treatment\_efficacy (q010)    & 0.20 & 0.40 \\
\midrule
\textbf{Average} & \textbf{0.32} & \textbf{0.28} \\
\bottomrule
\end{tabular}
\label{tab:adversarial}
\end{table}

This outcome is partially consistent with the experimental design: adversarial
documents carry source types with low reliability priors, and the
multiplicative penalty suppresses them in most cases when competing against
higher-reliability documents on the same topic. The exceptions at q006 and
q010 suggest that topic-specific corpus structure can create conditions where
the reranking interacts with document distribution in ways that are not
fully controlled by the prior alone. This should not be interpreted as
robustness against adversarial content in general---an adversary aware of
the prior system could assign high-reliability source labels to misleading
content, bypassing the penalty entirely. The contribution of this experiment
is to operationalize and measure the suppression effect under this specific
and limited threat model.

\begin{figure}[H]
  \centering
  \includegraphics[width=0.95\linewidth]{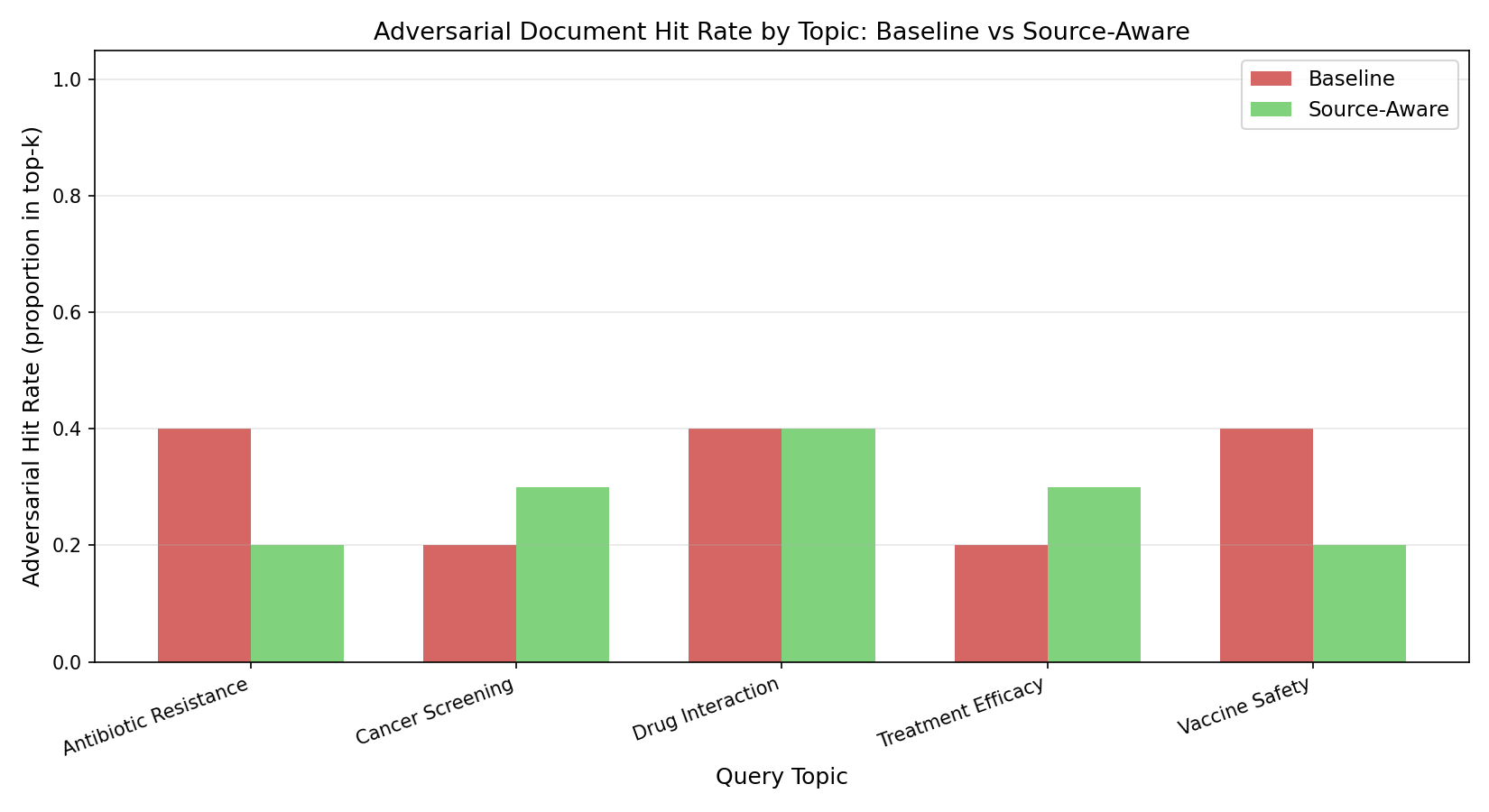}
  \caption{Adversarial document hit rate by query topic, baseline
           vs.\ source-aware. Under the evaluated threat model,
           source-aware reranking reduces average adversarial retrieval
           from 0.32 to 0.28, with mixed results across individual topics.}
  \label{fig:adversarial}
\end{figure}

\begin{figure}[H]
  \centering
  \includegraphics[width=0.95\linewidth]{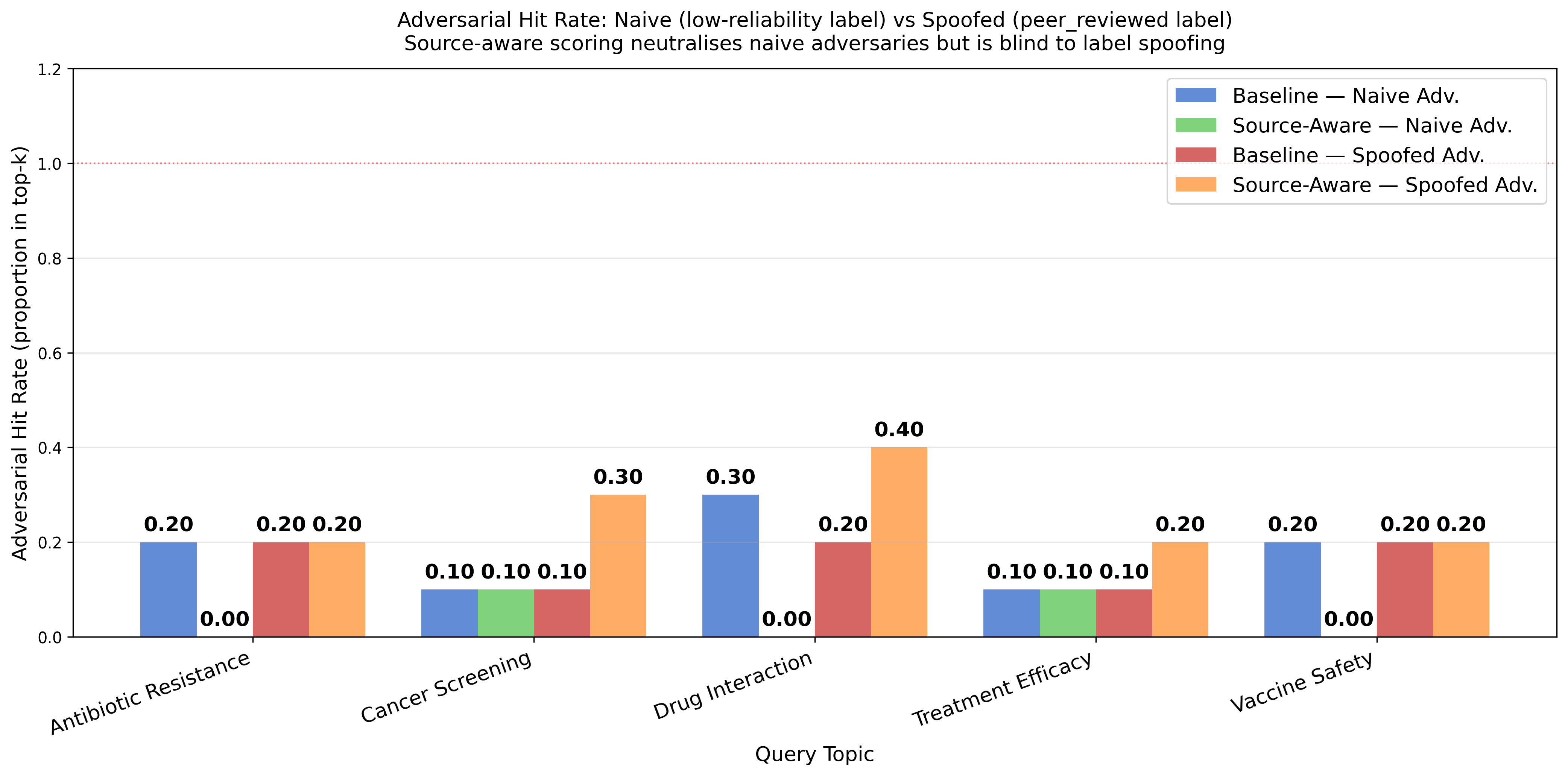}
  \caption{Spoofed vs.\ naive adversarial document retrieval rates across
           query topics. Spoofed documents, which carry artificially elevated
           source labels, are harder to suppress and expose the boundary
           conditions of the reliability prior approach.}
  \label{fig:spoofed}
\end{figure}

\subsection{Adaptive Weight Evolution}

Table~\ref{tab:adaptive} shows initial and final reliability weights after
10 sequential queries under oracle feedback simulation.

\begin{table}[H]
\centering
\caption{Adaptive weight evolution under oracle feedback simulation.}
\begin{tabular}{lccc}
\toprule
Source Type & Initial & Final & Delta \\
\midrule
peer\_reviewed       & 0.9500 & 1.0000 & $+$0.0500 \\
government           & 0.9000 & 0.9950 & $+$0.0950 \\
established\_news    & 0.7500 & 0.7750 & $+$0.0250 \\
organization\_report & 0.7000 & 0.7300 & $+$0.0300 \\
blog                 & 0.5000 & 0.5000 & 0.0000    \\
user\_generated      & 0.4000 & 0.4000 & 0.0000    \\
ai\_generated        & 0.3000 & 0.3000 & 0.0000    \\
unknown              & 0.3500 & 0.3500 & 0.0000    \\
\bottomrule
\end{tabular}
\label{tab:adaptive}
\end{table}

Government sources receive the largest positive update ($+$0.095),
followed by peer-reviewed sources ($+$0.050), organization reports
($+$0.030), and established news ($+$0.025), consistent with their
appearance in relevant top-5 results across queries.
Source types never retrieved in the top-5 receive no update---the
system correctly makes no inference about sources it has not observed.
The mean absolute weight change over the final three queries was below
the 0.01 convergence threshold, indicating stable behavior within 10
interactions. These results are produced under simulated oracle feedback
and reflect the theoretical behavior of the update rule rather than
evidence of deployed adaptive performance.

\begin{figure}[H]
  \centering
  \includegraphics[width=0.95\linewidth]{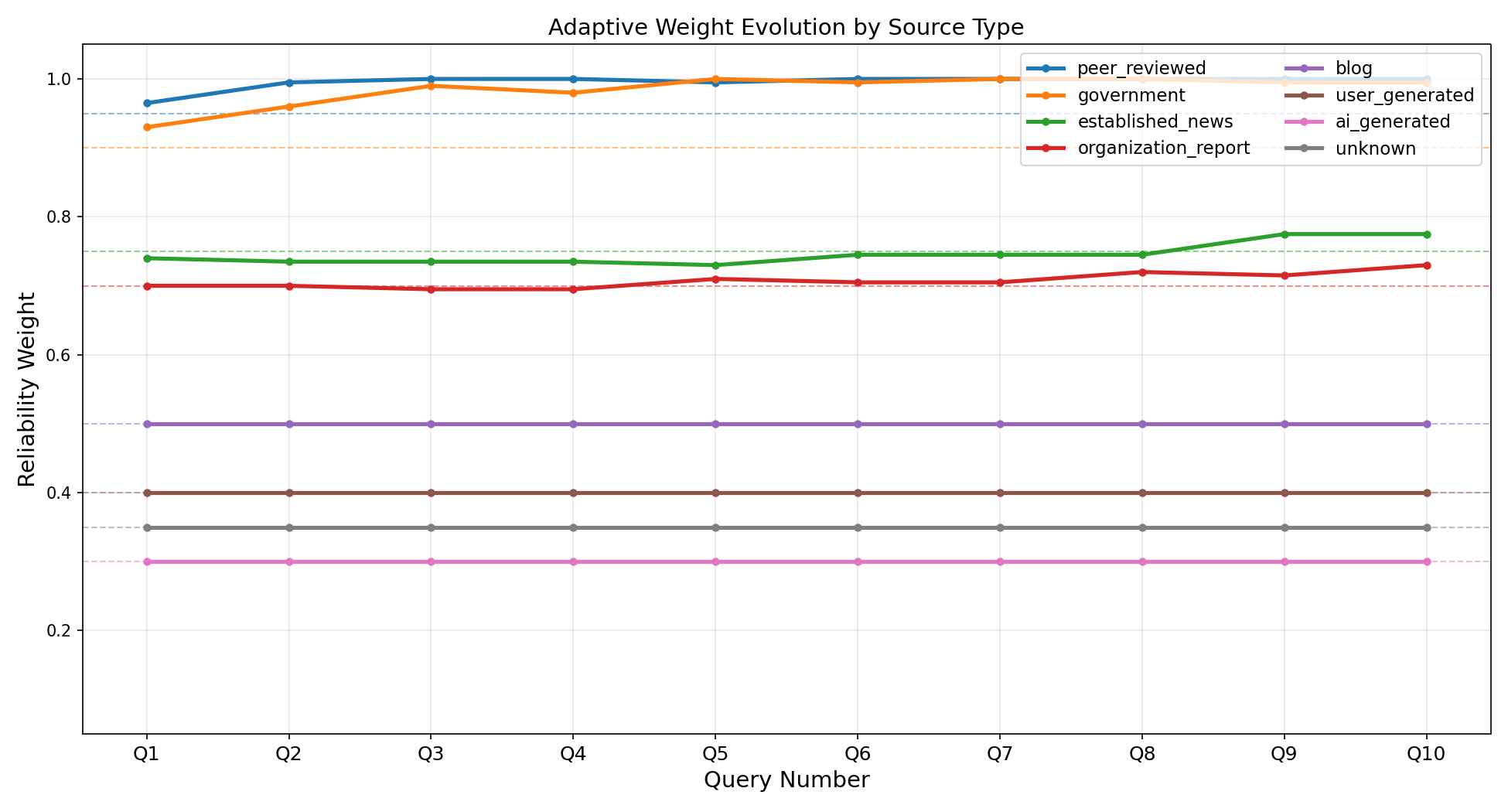}
  \caption{Reliability weight evolution across 10 sequential queries under
           oracle feedback simulation. Government and peer-reviewed sources
           receive the largest positive updates. Source types not retrieved
           in top-5 receive no update.}
  \label{fig:weightevo}
\end{figure}

\begin{figure}[H]
  \centering
  \includegraphics[width=0.95\linewidth]{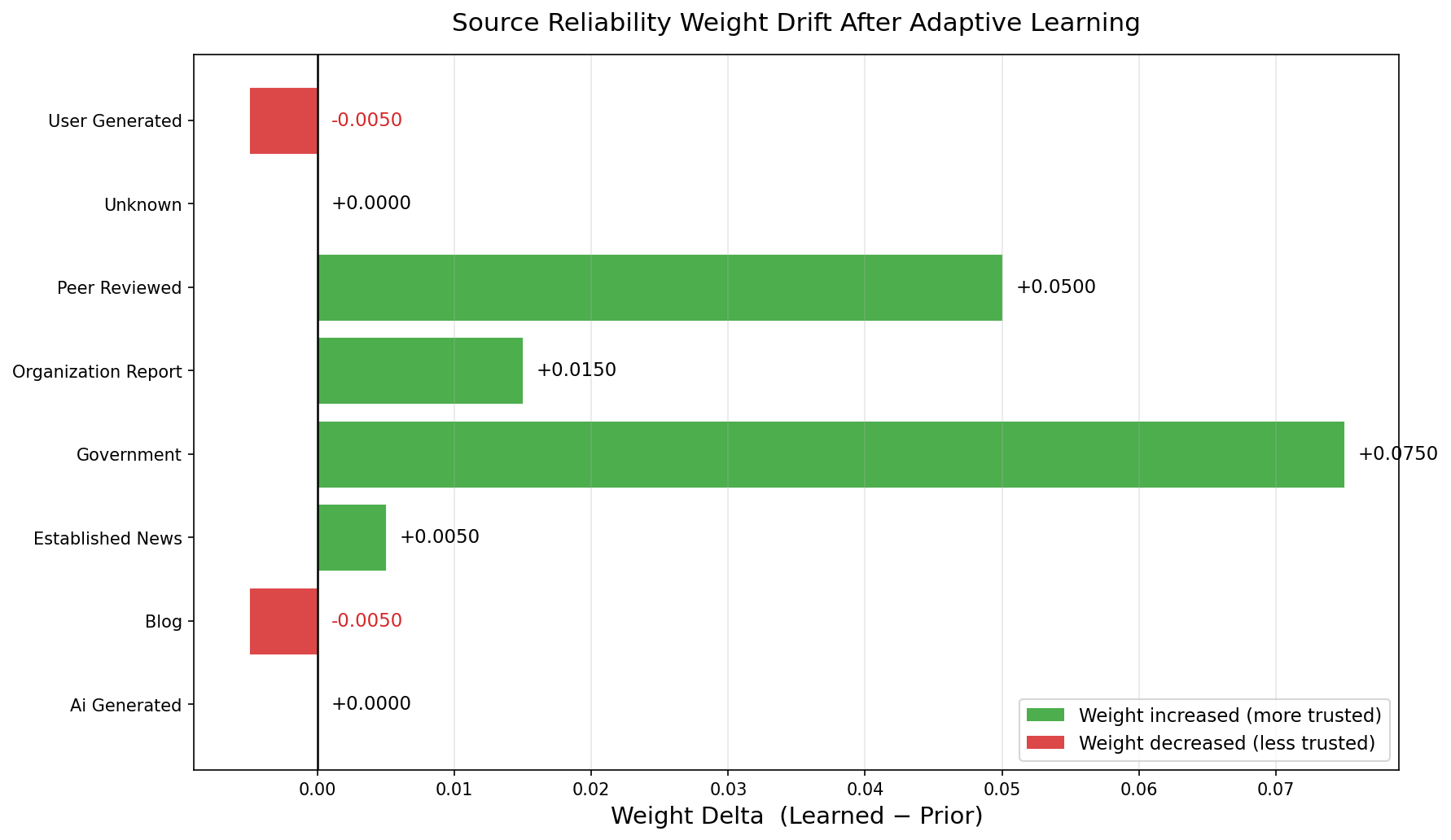}
  \caption{Change in reliability weight (final minus initial) per source
           type after 10 queries.}
  \label{fig:weightdelta}
\end{figure}

\subsection{Sensitivity Analysis}
\label{sec:sensitivity}

Table~\ref{tab:sensitivity} reports retrieval metrics across three weight
configurations evaluated on a held-out set of four queries (q007--q010).

\begin{table}[H]
\centering
\caption{Sensitivity analysis across weight configurations on test queries
         (q007--q010). Baseline values are identical across configurations
         as the baseline does not use reliability weights.}
\begin{tabular}{lcccc}
\toprule
Config & P@5 Baseline & P@5 SA & NDCG SA & MRR SA \\
\midrule
Compressed (spread 0.25) & 0.50 & 0.55 & 0.954 & 1.000 \\
Current (spread 0.65)    & 0.50 & 0.80 & 0.971 & 1.000 \\
Aggressive (spread 0.89) & 0.50 & 0.80 & 0.974 & 1.000 \\
\bottomrule
\end{tabular}
\label{tab:sensitivity}
\end{table}

The compressed configuration, which uses a narrower spread between high and
low reliability priors (0.25), produces only a modest improvement in
Precision@5 ($+$0.05). The current and aggressive configurations, with
spreads of 0.65 and 0.89 respectively, both achieve Precision@5 of 0.80
($+$0.30 over baseline). MRR is 1.000 across all three configurations,
indicating that the first retrieved result is always relevant regardless of
weight spread. The marginal difference between current and aggressive
configurations suggests diminishing returns beyond a spread of 0.65, and
that the current configuration represents a practical operating point. These
results confirm that prior spread is the primary driver of retrieval
improvement and that the current configuration is not a fragile choice.

\begin{figure}[H]
  \centering
  \includegraphics[width=0.95\linewidth]{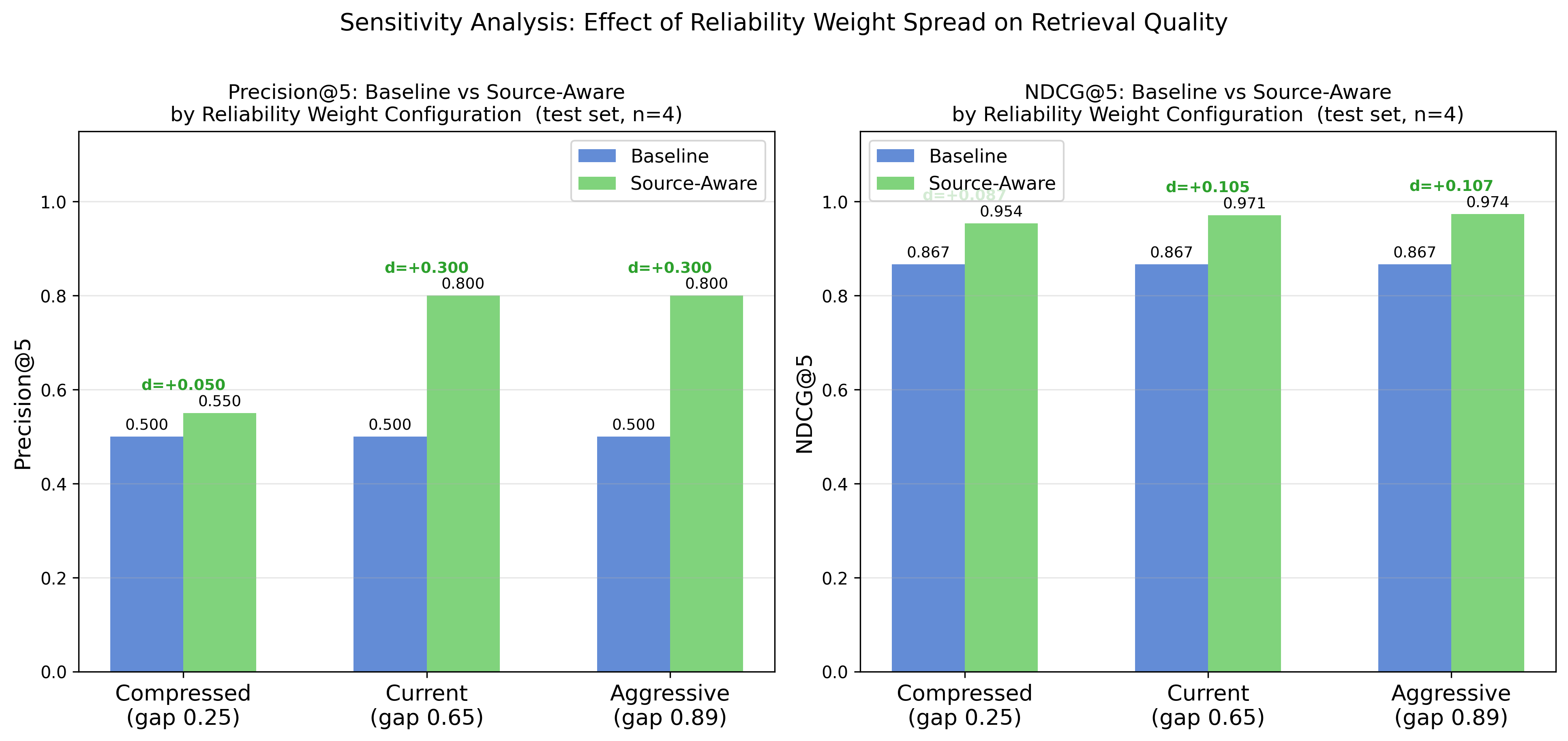}
  \caption{Precision@5 and NDCG@5 across compressed, current, and aggressive
           weight configurations on test queries (q007--q010). The current and
           aggressive configurations both outperform the compressed baseline
           by a substantial margin.}
  \label{fig:sensitivity}
\end{figure}

\section{Challenges}

The central limitation of this work is that reliability priors are manually
defined rather than learned from data. The values in Table~\ref{tab:priors}
represent reasonable domain assumptions but have not been empirically
validated against an independent credibility ground truth.

The adversarial evaluation uses explicitly labeled adversarial documents
assigned to low-reliability source types, meaning the method's partial
success in suppressing these documents is structurally expected given the
experimental design. A more rigorous adversarial evaluation would use
documents whose source type is unknown or misrepresented, which would
constitute a harder and more realistic threat model. The cases at q006 and
q010 where source-aware performed worse than baseline highlight the limits
of a purely prior-based defense and point toward the need for content-level
signals in future work.

The adaptive weight update relies on oracle feedback derived from
ground-truth relevance labels. This is not a deployable feedback mechanism.
Real deployment would require implicit feedback signals such as user
engagement, downstream answer quality, or expert annotation, none of which
were available in this study.

The corpus of 120 documents is domain-specific and small relative to
production RAG corpora. With only 10 evaluation queries, no formal
statistical significance testing was performed. Results from this scale
should be treated as exploratory and should not be extrapolated to larger or
multi-domain settings without further evaluation.

\section{Future Work}

Several directions follow from this work. Reliability priors could be
estimated from large-scale credibility datasets or learned jointly with the
retrieval model, removing the dependency on manual configuration. The
adaptive update mechanism could be extended to incorporate real implicit
feedback signals such as user dwell time or downstream answer quality from an
evaluator model. The single-hop neighbor smoothing could be extended to
multi-hop graph propagation, which may provide stronger reliability signal in
larger corpora where neighborhood structure is richer.

A more robust adversarial evaluation should test against a harder threat
model where adversaries can spoof source metadata, assigning high-reliability
labels to misleading content. The per-query variability observed in this
study, particularly the cases where source-aware retrieval performed worse
than baseline, suggests that a content-level credibility signal is needed to
complement the source-level prior.

Evaluation on established benchmarks such as BEIR would allow direct
comparison against prior retrieval systems across multiple domains and would
establish whether the observed improvements generalize beyond this controlled
setting. Scaling the corpus and query set would also enable formal
statistical significance testing, which was not possible at the current
evaluation scale.

\section{Conclusion}

This work introduced a source-aware reranking framework for
retrieval-augmented generation that assigns domain-informed reliability
priors to documents and reweights retrieval scores by source type. In a
controlled evaluation on a 120-document health-domain corpus executed on
Rosie, MSOE's high-performance computing cluster, the framework improves
Precision@5 from 0.48 to 0.72 and reduces average adversarial document
retrieval from 0.32 to 0.28 relative to a similarity-only baseline. MRR
improves to 1.000 across all queries, meaning the highest-ranked retrieved
document is always relevant. Sensitivity analysis confirms that the current
weight configuration is not a fragile choice, with both the current and
aggressive spread configurations producing consistent improvements over the
compressed baseline. These results suggest that lightweight source provenance
weighting is a viable and interpretable post-retrieval modification that does
not require changes to the underlying language model or embedding model. The
method's limitations---manual priors, oracle feedback, small-scale
evaluation, and a specific threat model---are acknowledged explicitly and
point to concrete directions for future work.

\bibliographystyle{plainnat}
\bibliography{references}

\end{document}